\documentclass[conference]{IEEEtran}
\IEEEoverridecommandlockouts
\usepackage{booktabs}
\usepackage{amsmath}
\usepackage{multirow}
\usepackage[numbers,sort&compress]{natbib}
\usepackage{amsmath,amssymb,amsfonts}
\usepackage{algorithmic}
\usepackage{graphicx}
\usepackage{textcomp}
\usepackage{xcolor}
\def\BibTeX{{\rm B\kern-.05em{\sc i\kern-.025em b}\kern-.08em
    T\kern-.1667em\lower.7ex\hbox{E}\kern-.125emX}}
\begin{document}





\title{ Enhancing Decision-Making in Optimization through LLM-Assisted Inference: A Neural Networks Perspective






\thanks{Gaurav Singh is with Institute for Infocomm Research (I²R), Agency for Science, Technology and Research (A*STAR), Singapore.  Kavitesh Kumar Bali with Centre for Frontier AI Research (CFAR), A*STAR, Singapore. \emph{Email:gaur1nov@gmail.com, bali.kavitesh@gmail.com}}
}

\author{\IEEEauthorblockN{Gaurav Singh}
\and
\IEEEauthorblockN{Kavitesh Kumar Bali}
}

\maketitle

\begin{abstract}

This paper explores the seamless integration of Generative AI (GenAI) and Evolutionary Algorithms (EAs) within the domain of large-scale multi-objective optimization. Focusing on the transformative role of Large Language Models (LLMs), our study investigates the potential of LLM-Assisted Inference to automate and enhance decision-making processes. Specifically, we highlight its effectiveness in illuminating key decision variables in evolutionarily optimized solutions while articulating contextual trade-offs. Tailored to address the challenges inherent in inferring complex multi-objective optimization solutions at scale, our approach emphasizes the adaptive nature of LLMs, allowing them to provide nuanced explanations and align their language with diverse stakeholder expertise levels and domain preferences. Empirical studies underscore the practical applicability and impact of LLM-Assisted Inference in real-world decision-making scenarios.

\end{abstract}


\section{Introduction}

In the evolving Artificial Intelligence (AI) landscape, the journey from traditional AI to Machine Learning (ML) and now Generative AI (GenAI) is noteworthy~\citep{stone2022artificial}. 
Initially, AI operated based on predefined rules to act on data, paving the way for more adaptive learning through ML. Now, with the advent of GenAI, we witness a transformative leap wherein the system not only learns from data but actively generates new data, unlocking unprecedented possibilities~\citep{WinNT}. 
\begin{figure}[htbp]
\includegraphics[width=\linewidth, height = 5.5cm]{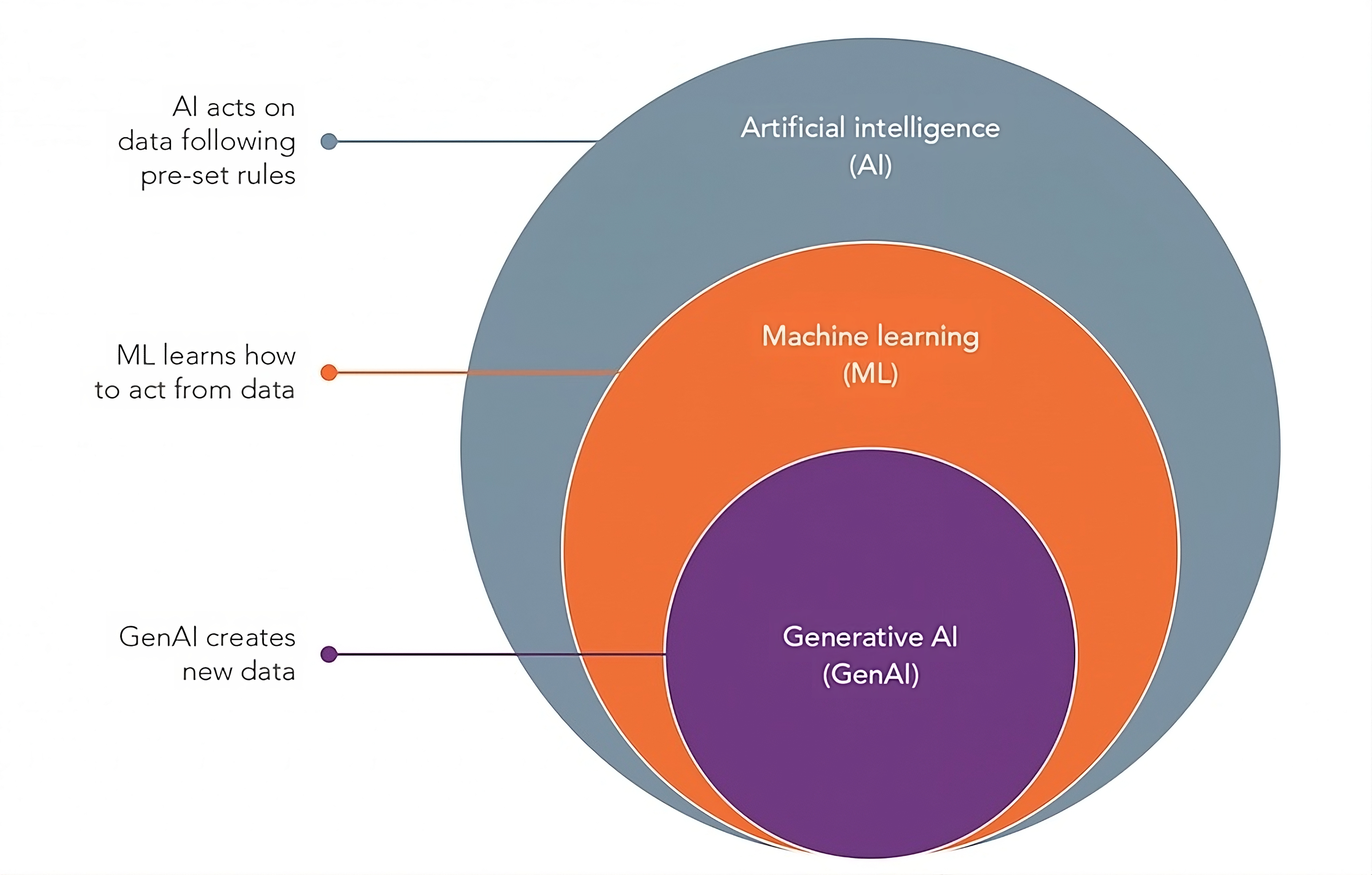}
\caption{Evolution: AI$\rightarrow$ ML$\rightarrow$ GenAI~\citep{WinNT}}
\label{fig_AGI}
\end{figure}


GenAI, encompassing models like GPT-3~\citep{brown2020language}, BERT~\citep{devlin2018bert}, DALL-E 2~\citep{ramesh2022hierarchical} and Imagen~\citep{saharia2022photorealistic} has witnessed unprecedented success in various domains, including natural language processing, image synthesis, and data analysis~\citep{cao2023comprehensive, wong2023prompt}. 
Unlike its predecessors, GenAI has the capacity to create novel content, transcending the limitations of rule-based AI and learned behaviors of ML (as per Fig~\ref{fig_AGI}). In this GenAI era, Large Language Models (LLMs) have emerged as transformative tools, showcasing remarkable language understanding and generation capabilities.

Concurrently, Evolutionary Algorithms (EAs) have proven to be robust tools in the vast domain of AI and optimization. To elaborate, EAs find application across multiple facets of ML: in preprocessing (e.g., feature selection and resampling), learning through neuroevolution (e.g., parameter tuning, membership functions, and neural network topology/autoML), and postprocessing (e.g., rule optimization, pruning decision trees/support vectors, and ensemble learning)~\citep{telikani2021evolutionary}.
Further, EAs play a crucial role in complex optimization settings, particularly in large-scale scenarios with numerous decision variables. Their implicit parallelism is noteworthy, demonstrating prowess in multi-objective settings by generating a diverse array of solutions in a single algorithmic run.




This paper explores the symbiotic relationship between Gen AI and EA-based optimization, with a specific focus on the role of LLMs in automating and enhancing decision-making processes. We introduce the concept of LLM-Assisted Inference, demonstrating how it streamlines the understanding and explanation of key decision factors, particularly in the context of inferring optimized solutions from evolutionary multi-objective optimization searches. The significance of LLMs becomes evident in large-scale, complex optimization scenarios, where numerous decision variables results in a diverse array of solutions projected on the Pareto front - a graphical representation highlighting trade-offs between conflicting objectives, offering a spectrum of optimal solutions in the context of evolutionary-driven optimization.

In the pursuit of \emph{"Explainable AI"}~\citep{linardatos2020explainable}, LLMs emerge as crucial players, serving as a bridge between technical complexities and comprehensibility across varying levels of expertise. From domain experts seeking detailed technical insights to high-level executives with limited technical knowledge, LLM-Assisted Inference enhances the interpretability of AI-generated solutions. This approach proves effective through a real-world study, demonstrating its practical use and impact in decision-making - as shall be showcased in Section~\ref{sec:casestudy}\footnote{Note: This seminal paper emphasizes LLM-based inference, prioritizing it over exhaustive details of EA-driven multi-objective optimization due to space constraints and focused objectives. The intent is to highlight  the significant role of LLMs in enhancing decision-making within optimization scenarios.}. 

\section {Large Language Models}
\subsection{Evolution and Development}

LLMs like the GPT (Generative Pre-trained Transformer) series have been pivotal in advancing natural language processing. The GPT series began with the original GPT and evolved through GPT-2, GPT-3, and the latest GPT-4, each surpassing its predecessor in complexity and capabilities.

GPT-2, introduced by Radford et al. (2019), marked a significant advancement, demonstrating the ability to generate coherent and contextually relevant text over extended passages \citep{radford2019language}. GPT-3, developed by Brown et al. (2020), scaled up to 175 billion parameters and broadened the scope of linguistic tasks without task-specific training \citep{brown2020language}. GPT-4 continues this evolution with even greater language understanding and generation capabilities \citep{bubeck2023sparks}.

\subsection{Applications}

LLMs have proven their versatility across multiple use cases. Their applications range from text generation, language translation, and content creation to complex tasks like coding assistance, data analysis, and decision support \citep{applicationsLLM}. They have also been applied in sectors like healthcare for medical documentation, in legal for document drafting and review, and in education for personalized learning experiences \citep{healthcareApplication, legalApplication, educationApplication}.

\section{Evolutionary Multi-objective Optimization: A brief review} 
Evolutionary Multi-objective Optimization (EMO) is a powerful paradigm within the field of optimization/transfer-optimization~\citep{bali2020cognizant}, specifically designed to handle problems with multiple conflicting objectives. Unlike traditional single-objective optimization, EMO aims to find a set of solutions that represents a trade-off among different objectives, known as the Pareto front~\citep{coello2006evolutionary}.

\subsection{Basic Formulations and Definitions:}
In EMO, a multi-objective optimization problem is typically defined as follows. Given a vector of decision variables $\mathbf{x} = [x_1, x_2, ..., x_n]$, and a vector of objective functions $\mathbf{f} = [f_1(\mathbf{x}), f_2(\mathbf{x}), ..., f_m(\mathbf{x})]$, where $m$ is the number of objectives, the goal is to find a set of solutions $\mathbf{X}$ that minimizes or maximizes each objective simultaneously.

\textbf{Pareto Front:}
The solutions in $\mathbf{X}$ are evaluated based on the Pareto dominance concept. A solution $\mathbf{x}_1$ is said to dominate another solution $\mathbf{x}_2$ if $\forall i, f_i(\mathbf{x}_1) \leq f_i(\mathbf{x}_2)$ and $\exists j, f_j(\mathbf{x}_1) < f_j(\mathbf{x}_2)$. The set of non-dominated solutions forms the Pareto front, representing optimal trade-offs between conflicting objectives.

\textbf{Visualization:}
Below is a simple illustration of a 2-objective Pareto front for visualization:

\begin{figure}[htbp]
    \centering
    \includegraphics[width=0.6\linewidth]{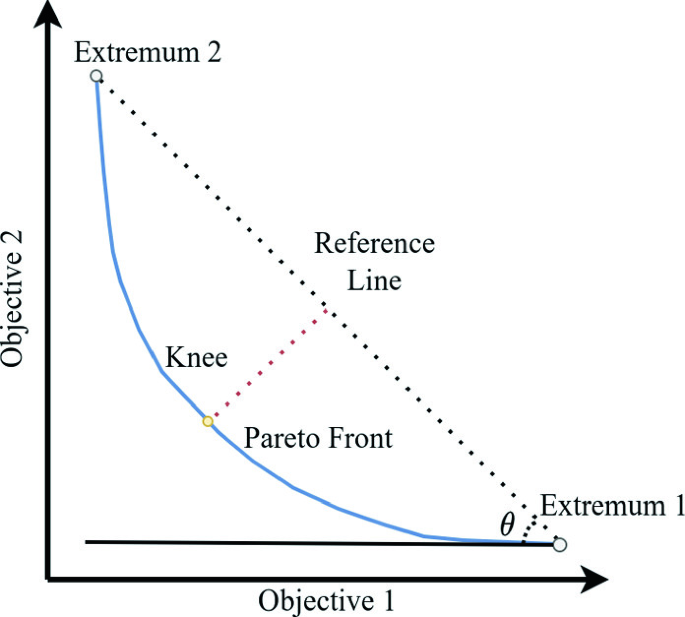}
    \caption{Visualization of a 2-objective Pareto front for a minimization problem generated by an evolutionary algorithm. Solutions at the extremes prioritize a single objective, while those in the middle characterize knee solutions, achieving a balanced compromise between conflicting objectives~\citep{heidari2022finding}.}
    \label{fig:pareto_front_example}
\end{figure}

This figure depicts a trade-off between two conflicting objectives, and the solutions along the Pareto front represent the optimal compromise between these objectives. Popular algorithms such as NSGA-II (Non-dominated Sorting Genetic Algorithm II) and MOEA/D (Multi-Objective Evolutionary Algorithm based on Decomposition) are commonly employed to generate such Pareto fronts in evolutionary multi-objective optimization scenarios~\citep{li2008multiobjective}.

\section {Explainable AI}

\subsection{Brief Review of Explainable AI}

Explainable AI (XAI) has become a pivotal element in artificial intelligence systems, aiming to enhance the transparency and interpretability of complex models~\citep{linardatos2020explainable}~\citep{lundberg2017unified}. The demand for understanding the decision-making processes of AI systems has grown significantly, especially in domains where decisions have substantial impacts.

Explainable AI techniques vary from model-agnostic approaches, such as LIME (Local Interpretable Model-agnostic Explanations)~\citep{ribeiro2016should}, to model-specific methods like SHAP (SHapley Additive exPlanations)~\citep{lundberg2017unified}. These techniques aim to provide insights into the inner workings of machine learning models, making them more accessible and trustworthy.

\subsection{Connecting LLMs for Explainability}

Large Language Models (LLMs) play a crucial role in the realm of Explainable AI. Their capacity for natural language processing allows them to articulate complex model behaviors and predictions in a human-understandable manner. LLMs facilitate the interpretation and explanation of AI models, making them more accessible to a wider audience.

In the context of this paper, LLMs are particularly instrumental in enhancing and automating the process of explaining optimized solutions resulting from evolutionary multi-objective optimization. The subsequent section showcases how LLM-Assisted Inference not only aids in interpreting large sets of optimized solutions but also caters to decision-makers at various levels, providing nuanced insights based on their expertise and requirements.

\section{LLM-Assisted inference: A case study}
\label{sec:casestudy}
In this section, we explore the application of LLM-Assisted Inference in a complex multi-objective optimization problem within a sustainability production environment. The case study introduces a challenging optimization problem aimed at achieving crucial sustainability objectives. A vivid illustration of trade-offs between the two objectives is provided through a Pareto front generated by a candidate Evolutionary Algorithm (EA). The exploration of LLM-Assisted Inference is multifaceted, encompassing various perspectives. Automated inferencing filters and highlights decision variables with the most significant impact on achieving optimized solutions. Additionally, the process includes solution interpretation catering to different levels of expertise and goals, providing insights for both technical experts and high-level decision-makers.

This case study serves as a practical demonstration of how LLM-Assisted Inference can enhance the understanding and interpretation of solutions within the context of complex optimization, contributing to more informed decision-making in sustainability-focused production environments.

\subsection{Case study overview: Sustainable Infrastructure Planning via EMO}

 Achieving Sustainable Infrastructure Development: 
 The infrastructure project optimization pertains to a wide range of infrastructure types, encompassing sectors such as transportation (e.g., roads, bridges, public transit), energy (e.g., renewable energy installations, power grids), urban development (e.g., housing, utilities), and environmental projects (e.g., water management, conservation efforts). This problem is of utmost relevance in a world where infrastructure development impacts society, the economy, and the environment, making it essential to optimize resource allocation for long-term well-being.  A multi-Objective Optimization Approach is often adopted to address the challenge of developing a sustainable infrastructure plan that balances economic, environmental, and social objectives across numerous decision variables on a large scale. In this case, the multi-objective optimization framework applies to diverse infrastructure categories where the trade-offs among cost, environmental impact, efficiency, durability, innovation, and energy usage are critical considerations for achieving sustainable development.

 \subsubsection{Optimization Problem}
 In this paper, we consider a complex multi-objective Sustainable Infrastructure planning problem inspired by \cite{ADSHEAD2019101975,9345531, RAEI2019124091}. 
 The optimization problem at hand involves a substantial number of decision variables, reaching up to 50. A selection of decision variables is outlined in Table~\ref{table:decision_variables_full}. The array of decision variables reflects a comprehensive consideration of economic, environmental, and technical aspects, aligning with a holistic approach to sustainable infrastructure planning.

 \begin{table}[htbp]
\scriptsize
\begin{tabular}{|c|l|}
\hline
\textbf{Variable Number} & \textbf{Decision Variable Description} \\
\hline
1 & Cost Efficiency: Output per dollar spent \\
\hline
2 & Durability: Expected lifespan of the infrastructure \\
\hline
3 & Renewable Energy Usage: Percentage of energy from renewable sources \\
\hline
4 & Carbon Footprint: Total greenhouse gas emissions \\
\hline
5 & Water Usage: Total annual water consumption \\
\hline
6 & Waste Production: Total waste generated annually \\
\hline
7 & Land Use: Total land area required \\
\hline
8 & Energy Efficiency: Overall energy efficiency of the project \\
\hline
9 & Maintenance Cost: Annual cost for maintenance \\
\hline
10 & Innovation Index: Score for technological innovation \\
\hline
... & ... \\
\hline
50 & Supply Chain Stability: Reliability and stability of the supply chain \\
\hline
\end{tabular}
\caption{List of All Decision Variables in Sustainable Infrastructure Planning. Due to space constraints, a limited subset of decision variables is showcased in this table.}
\label{table:decision_variables_full}
\end{table}

The objective functions for the optimization problem are defined as follows: 

 \begin{enumerate}
    \item \textbf{Objective 1: Minimize Total Cost}
    \begin{equation}
        \text{Total Cost} = \sum_{i=1}^{n} \text{Cost}_{i}
    \end{equation}
    Where $\text{Cost}_{i}$ represents individual cost components like construction, maintenance, and operational costs \footnote{All costs are normalized by a factor of million i.e., million dollars}.
    \item \textbf{Objective 2: Minimize Environmental Impact}
    \begin{equation}
        \text{Environmental Impact} = \sum_{j=1}^{m} w_j \times \text{Impact}_{j}
    \end{equation}
    Where $w_j$ are the weights and $\text{Impact}_{j}$ are individual environmental impact factors like emissions, waste production, and land use\footnote{Given the extensive number of decision variables (50), the intricate details of the problem formulation have been intentionally withheld in this paper to streamline the focus on the broader application of LLMs in optimization scenarios}. 
    
\end{enumerate}


\begin{table*}[htbp]
\centering
\begin{tabular}{|c|c|c|c|c|c|c|}
\hline
\textbf{Sol. \#} & \textbf{Total Cost (M\$)} & \textbf{Env. Impact (Score)} & \textbf{Cost Effic. (Units/\$)} & \textbf{Durability (Years)} & \textbf{Renewable Energy (\%)} & \textbf{... (upto 50 Vars)} \\
\hline
1 & 200.00 & 1.004 & 50 & 25 & 15 & ... \\
\hline
51 & 202.00 & 0.910 & 49 & 27 & 18 & ... \\
\hline
101 & 204.00 & 0.807 & 48 & 29 & 20 & ... \\
\hline
201 & 208.01 & 0.709 & 46 & 32 & 25 & ... \\
\hline
301 & 212.01 & 0.573 & 44 & 35 & 30 & ... \\
\hline
401 & 216.02 & 0.463 & 42 & 38 & 35 & ... \\
\hline
... & ... & ... & ... & ... & ... & ... \\
\hline
500 & 219.98 & 0.328 & 40 & 40 & 40 & ... \\

\hline
\end{tabular}
\caption{Selected Pareto Optimal Solutions with a subset of Decision Variables}
\label{table:pareto_solutions_selected}
\end{table*}



\subsubsection{EMO-Driven Optimization Approach}

To address the outlined multi-objective problem, we employ an EMO approach utilizing a candidate evolutionary algorithm for optimization. Specifically, NSGA-II is chosen to tackle the problem, with a population size of 500. Due to space constraints, intricate algorithmic and optimization details are omitted from this paper\footnote{Note: The primary focus of this paper is on LLM-assisted inference rather than evolutionary optimization. It emphasizes the pivotal role of LLMs in enhancing solution inferencing at scale and improving decision-making processes in large-scale multi-objective optimization settings.
}.

\begin{figure}[ht]
\centering
\includegraphics[width=\linewidth]{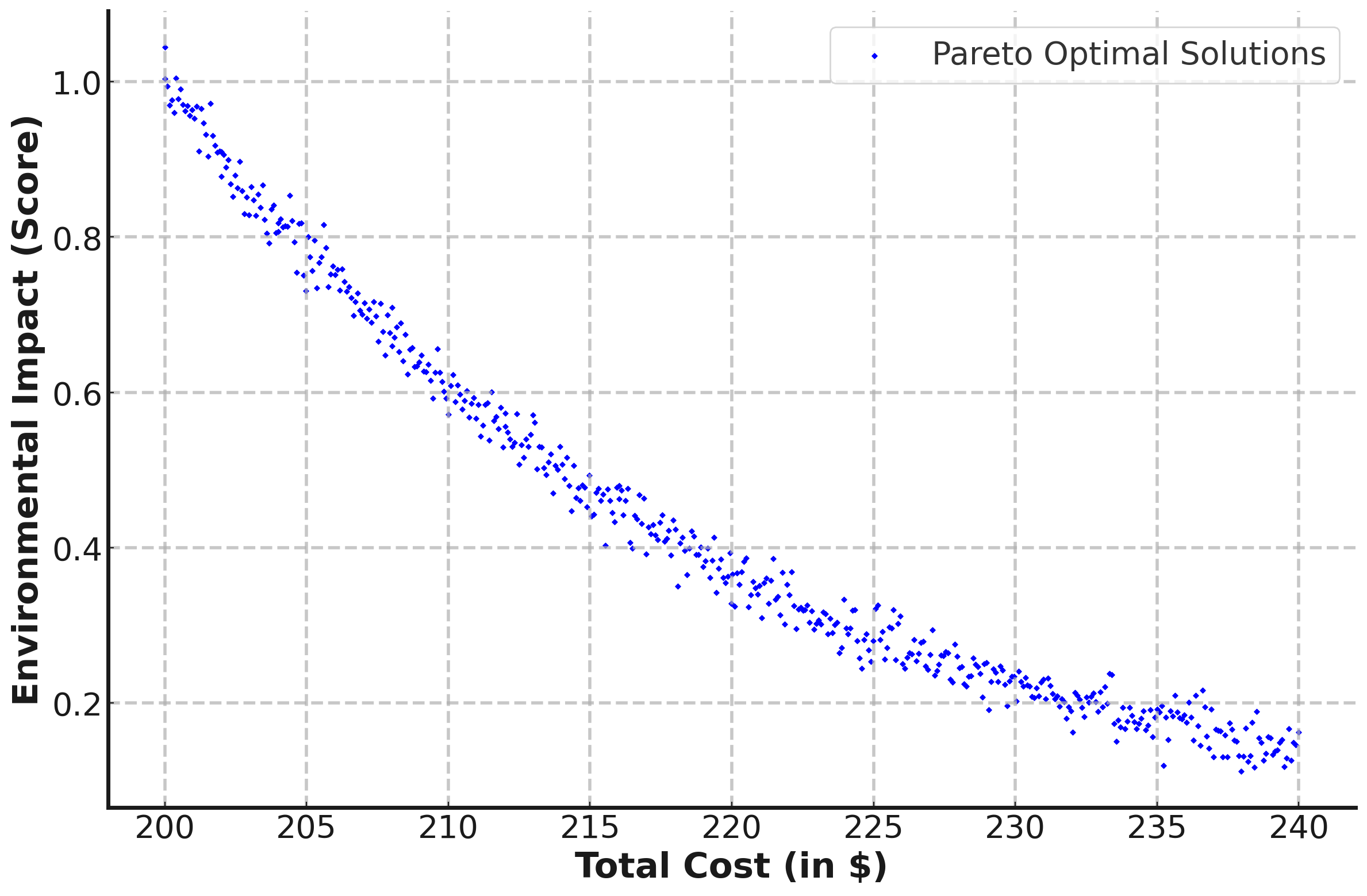}
\caption{Pareto Front Illustrating the Trade-off Between Total Cost (M\$) and Environmental Impact}
\label{fig:pareto_front}
\end{figure}

A snippet of the obtained optimized solutions is presented in Table~\ref{table:pareto_solutions_selected}. Correspondingly, the image of optimal solutions in the objective space is depicted by the Pareto front in Figure~\ref{fig:pareto_front}.

\subsection{Standard Approach to Solution Inferencing: Manual Inspection and Statistical Analysis}

In traditional practices, decision-makers commonly resort to manual inspection of data, such as sets of optimized solutions, and employ standard statistical analyses, like correlation coefficients, to unveil patterns and relationships between decision variables and objectives. While these methods offer insights, they often prove inadequate in providing nuanced and comprehensive understanding, especially in complex multi-objective optimization scenarios. This limitation becomes more noticeable, particularly when dealing with a large number of decision variables.


\subsection{LLM-Assisted Inference}
In response to the limitations of conventional approaches, this paper advocates for LLM-assisted inference to enhance and automate decision-making in large-scale, complex settings. The paper explores the value of LLMs from various perspectives, including the filtering of important decision variables, inferencing tailored for different levels of expertise ranging from technical decision-makers to high-level executives, and inferencing aligned with diverse goals and requirements.

The subsequent sections present detailed analytical results, providing empirical evidence to substantiate the effectiveness of LLM-assisted inference within the context of sustainable infrastructure planning. It's essential to note that the underlying LLM model employed was GPT-3.5, interfaced through the ChatGPT UI tool\footnote{The choice of GPT-3.5 as the baseline model is rooted in its open-source nature, promoting accessibility for researchers to actively engage in and contribute to such explorative studies.}. \\

\subsubsection{Filtering of Important Decision Variables}
LLMs demonstrate the ability to identify and filter out the most influential decision variables crucial for decision-making. An example of a corresponding prompt used to query the LLM is as follows:
\begin{itemize}
    \item \textbf{Prompt:} "Categorize decision variables in terms of their importance in the Pareto solutions."
\end{itemize}

\begin{table}[htbp]
\centering
\begin{tabular}{|l|l|}
\hline
\textbf{Variable Type} & \textbf{Decision Variables} \\
\hline
Primary & Renewable Energy Usage, Cost Efficiency, Durability, \\
& Carbon Footprint \\
\hline
Secondary & Water Usage, Waste Production, Land Use, \\
& Maintenance Cost, Energy Efficiency \\
\hline
Additional & Innovation Index, Environmental Impact Score, \\
& Community Impact, Regulatory Compliance, \\
& Stakeholder Satisfaction \\
\hline
\end{tabular}
\caption{Categorization of Decision Variables in Sustainable Infrastructure Planning}
\label{table:decision_variables_revised_categorization}
\end{table}

A snippet of a subset of filtered decision variables is showcased in Table~\ref{table:decision_variables_revised_categorization}.
As observed, it provides a categorization of decision variables in sustainable infrastructure planning. The variables are grouped into Primary, Secondary, and Additional categories, each contributing distinct factors such as Cost Efficiency, Environmental Impact Score, Innovation Index, and more. 
Moreover, a sample LLM-assisted inference considering the important variables is provided below:

\begin{itemize}
  \item \textbf{Prompt: } "Analyze key trade-offs between cost, durability, and renewable energy in our Pareto optimal solutions to identify the most balanced infrastructure project."
  \item \textbf{Focus on Key Decision Variables:}
    \begin{itemize}
      \item Analyzing Solution 51 with a focus on Cost Efficiency, Renewable Energy Usage, and Durability.
    \end{itemize}
  \item \textbf{Inference from Solution 51:}
    \begin{itemize}
      \item Total Cost: 202.00 M\$, Environmental Impact: 0.910.
      \item Cost Efficiency: 49 Units/\$, Durability: 27 Years, Renewable Energy Usage: 18\%.
      \item Compared to the extreme solution with the lowest total cost (Total Cost: 200.00 M\$, Env. Impact: 1.004), a slight increase in renewable energy usage in Solution 51 contributes to a reduction in the environmental impact score.
      \item The balance between durability and cost efficiency suggests strategic material and process selection.
    \end{itemize}
  \item \textbf{Justification:}
    \begin{itemize}
      \item Decrease in Environmental Impact Score from 1.004 (in the extreme low-cost solution) to 0.910 in Solution 51, with a minimal cost increase.
      \item Demonstrates how targeted adjustments in key variables can enhance environmental sustainability without significantly increasing costs. \\
    \end{itemize}
\end{itemize}

\subsubsection{Automated Inferencing of Pareto Front at Scale via LLMs}

LLMs demonstrate the capability to efficiently and automatically infer a large number of optimized solutions. As presented in this section, the analysis of the Pareto front (Fig.~\ref{fig:pareto_front}) for sustainable infrastructure planning has unveiled key solutions, each signifying a distinctive trade-off between total cost and environmental impact. Solutions at the extremes prioritize a single objective, while those in the middle, known as knee solutions, achieve a balanced compromise between conflicting objectives. Examples of such solutions include:
\begin{itemize}
    \item \textbf{Extreme Solution with the Lowest Total Cost:}
    \begin{itemize}
        \item Total Cost: 200 M\$
        \item Environmental Impact: $\approx$ 1.004
        \item Inference: Most cost-effective option with high environmental impact.
    \end{itemize}
    \item \textbf{Extreme Solution with the Highest Total Cost:}
    \begin{itemize}
        \item Total Cost: 240 M\$
        \item Environmental Impact: $\approx$ 0.115
        \item Inference: Highest investment for lowest environmental impact.
    \end{itemize}
    \item \textbf{Knee Solution:}
    \begin{itemize}
        \item Total Cost: $\approx$ 218.66 M\$
        \item Environmental Impact: $\approx$ 0.401
        \item Inference: Balanced trade-off between cost and environmental impact.
    \end{itemize}
    \item \textbf{Randomly Selected Solution:}
    \begin{itemize}
        \item Total Cost: $\approx$ 204.00 M\$
        \item Environmental Impact: $\approx$ 0.807
        \item Inference: Moderate cost and environmental impact. \\
    \end{itemize}
\end{itemize} 

\subsubsection{Inferencing for Different Levels of Domain Expertise}

This section explores the adaptability of LLM-assisted inference to different levels of domain expertise. The exploration includes tailored explanations ranging from technical details for domain experts to high-level insights for decision-makers with limited technical knowledge. To illustrate, LLM-assisted inference is applied to randomly selected solutions, providing nuanced explanations suited to various levels of expertise: 

\subsection*{For Domain Experts}
\begin{itemize}
  \item \textbf{Renewable Energy Usage:}
    \begin{itemize}
      \item Solution 404 shows a significant increase in renewable energy usage from 15\% to 55\%.
      \item This adjustment leads to a reduction in Environmental Impact Score by over 54\%.
      \item Highlights the correlation between renewable energy integration and environmental performance.
    \end{itemize}
\end{itemize}
\subsection*{For Mid-Level Technical Staff}
\begin{itemize}
  \item \textbf{Cost Efficiency:}
    \begin{itemize}
      \item Solution 232 shows an increase in cost efficiency to 46 Units/\$.
      \item Results in a better Environmental Impact Score, improving from 1.004 to 0.709.
      \item Demonstrates the effectiveness of investing in better cost efficiency.
    \end{itemize}
\end{itemize}
\subsection*{For Decision-Makers (Non-Technical)}
\begin{itemize}
  \item \textbf{Durability:}
    \begin{itemize}
      \item Increasing Durability from 25 to 35 years leads to a lower Environmental Impact Score.
      \item Balances the increase in Total Cost with long-term environmental benefits.
      \item Strategic consideration for long-term sustainability.\\
    \end{itemize}
\end{itemize}

\subsubsection{Inferencing for Different Goals/Requirements}
These solutions provide a spectrum of options, 
aiding decision-makers in selecting the most suitable approach for their specific goals and constraints. 

\begin{itemize}
  \item \textbf{Environmental Goals (for Environmental Groups):}
    \begin{itemize}
      \item Solution 57: Our project, with its lower environmental impact score of 0.910, significantly reduces greenhouse gas emissions.
      \item Solution 339: By allocating 30\% of our energy from renewable sources (shown in solution 339 with an impact of 0.573), we enhance sustainability.
      \item These efforts align our project with environmental goals and demonstrate a commitment to reducing ecological impact.
    \end{itemize}
  
  \item \textbf{Financial Viability (for Investors):}
    \begin{itemize}
      \item Solution 189: With a cost efficiency of 48 units per dollar, our project (solution 189) ensures robust financial returns.
      \item Solution 206: The project's long-term viability is further evidenced by the annual maintenance cost of only 10 M\$, as seen in solution 206.
      \item These factors make our project a financially attractive opportunity for investors.
    \end{itemize}
  
  \item \textbf{Community-Friendly Design (for Local Communities):}
    \begin{itemize}
      \item Solution 441: Demonstrating a community-friendly approach, solution 441 requires 20\% less land than typical projects, minimizing local disruption.
      \item In addition, the lower noise levels associated with this solution respect the community's need for a peaceful environment.
    \end{itemize}
  
  \item \textbf{Regulatory Compliance (for Government Agencies):}
    \begin{itemize}
      \item Solution 308: Our commitment to regulatory compliance is highlighted in solution 308, where we achieve a 100\% compliance score.
      \item The project's impeccable safety record further assures government agencies of our adherence to the highest standards.
    \end{itemize}
  
  \item \textbf{Socioeconomic Impact (for Broader Societal Goals):}
    \begin{itemize}
      \item Solution 470: The generation of 500 jobs, as demonstrated in solution 500, indicates a significant positive impact on local employment.
      \item The project's substantial contribution to the local economy, evidenced by the economic data in this solution, benefits the broader society.
    \end{itemize}
\end{itemize}

\subsection{Value Added by LLMs}
\subsubsection{Nuanced Explanation}

LLMs extend beyond numerical data to provide in-depth, nuanced explanations, enhancing the understanding of complex trade-offs in scenarios like task prioritization and resource allocation. For instance, in our case study, the LLM explained the implications of increasing Durability on both the project's long-term sustainability and initial costs, offering a deeper understanding beyond just the numerical increase in cost.

\subsubsection{Contextual Trade-offs}

LLMs excel in articulating inherent trade-offs in solutions, offering contextual insights crucial for informed decision-making. In the sustainable infrastructure planning case study, this was evident when the LLM highlighted how adjusting Renewable Energy Usage from 15\% to 55\% significantly reduced the Environmental Impact Score, contextualizing the environmental benefits against the cost implications.

\subsubsection{User-Friendly Descriptions}

LLMs generate descriptions that are accessible to a wide range of technical expertise, making complex optimization results understandable and actionable. For example, for non-technical decision-makers in our case study, the LLM simplified the trade-offs between cost and environmental impact, making the complex data from the Pareto front more interpretable.

\subsubsection{Handling Complexity}

In scenarios with numerous decision variables, LLMs decipher complexities and clarify how each decision contributes to overall trade-offs. This was demonstrated in our case study when the LLM adeptly navigated the complex relationship between Cost Efficiency and Environmental Impact, explaining how slight adjustments could lead to substantial environmental benefits.

\subsubsection{Adaptability to Stakeholder Language}

LLMs adapt their language to match stakeholders' domain-specific terminology, ensuring that interpretations resonate across various roles and expertise levels. In the case study, this adaptability was showcased by tailoring explanations for different stakeholders, using specific industry terms for domain experts while simplifying the language for local community representatives to ensure clarity and engagement.

\subsection{The Edge of LLMs: A Summary}

By leveraging LLM-assisted inference, decision-makers can gain profound insights into the characteristics of individual solutions on the Pareto front. This facilitates a more informed, nuanced, and sophisticated decision-making process compared to traditional approaches. The contextual, user-friendly, and adaptive nature of LLM interpretations adds a layer of sophistication, enhancing the overall interpretability and effectiveness of decision support in complex optimization scenarios.


\section{conclusion} 

This paper explores the symbiotic relationship between Gen AI and EAs, highlighting the pivotal role of LLMs in automating and enhancing decision-making processes. In the context of EMO, we leverage optimized solutions as an example to introduce LLM-Assisted Inference. This transformative approach aims to enhance our understanding and explanation of key decision variables in complex optimization scenarios.

The paper showcases the edge of LLMs over standard approaches, emphasizing their ability to provide nuanced explanations, articulate contextual trade-offs, and generate user-friendly descriptions. LLMs excel in handling the complexity inherent in large-scale multi-objective optimization, adapting their language to align with stakeholders' expertise levels and domain preferences.

Through a case study in Sustainable Infrastructure Planning, we demonstrate LLM-based inferencing, including the filtering of important decision variables, the automated analysis of the Pareto front, and tailored inferencing for different levels of expertise. The results substantiate the efficacy of LLM-Assisted Inference in providing meaningful insights, enhancing transparency, and facilitating informed decision-making.

In essence, this paper contributes to the evolving landscape of AI-driven optimization, emphasizing the valuable role of LLMs in augmenting explainability and interpretability, ultimately fostering more informed and impactful decision-making processes.

\bibliographystyle{IEEEtran}
 {\bibliography{main}}   

\end{document}